\documentclass[runningheads,a4paper]{llncs}

\usepackage{amssymb}
\usepackage{graphicx}

\usepackage{graphicx}
\usepackage[ruled,noend]{algorithm2e}
\usepackage{algpseudocode}
\usepackage{graphicx}
\usepackage{multirow}
\usepackage{float}
\usepackage[outerbars]{changebar}
\usepackage{amssymb,amsmath}
\usepackage{dsfont}

\usepackage{listings}
\usepackage[pagebackref=false,pdfborder={0, 0, 0},pdftex, bookmarks=2]{hyperref}
\hypersetup{
    pdftitle={Text Classification: A Sequential Reading Approach},
    colorlinks,%
    citecolor=black,%
    filecolor=black,%
    linkcolor=black,%
    urlcolor=black
}

\DeclareMathOperator*{\argmax}{argmax}
\DeclareMathOperator*{\argmin}{argmin}
\newcommand{\documents}{\mathcal{D}}

\newcommand{\doc}[1]{d_{#1}}
\newcommand{\documentstrain}{\mathcal{D}_{train}}

\newcommand{\nbdocumentstrain}{N_{train}}

\newcommand{\nbcategories}{C}
\newcommand{\categories}{\mathcal{Y}}
\newcommand{\scorecategorydocument}[1]{y^{#1}}
\newcommand{\scorecategorydocumentcategory}[2]{y^{#1}_{#2}}
\newcommand{\estimatedscorecategorydocumentcategory}[1]{\hat{y}_{#1}}
\newcommand{\sentence}[2]{\delta^{#1}_{#2}}
\newcommand{\lengthdoc}[1]{n_{#1}}

\newcommand{\states}{\mathcal{S}}
\newcommand{\actions}{\mathcal{A}}
\newcommand{\possibleactions}[1]{\mathcal{A}(#1)}
\newcommand{\reward}{r}
\newcommand{\rewardsa}[2]{r(#1,#2)}
\newcommand{\transitions}{T}
\newcommand{\transition}[2]{T(#1,\textrm{\textit{#2}})}

\usepackage{url}

\begin{document}

\mainmatter  

\title{Text Classification: A Sequential Reading Approach}

\author{Gabriel Dulac-Arnold \and Ludovic Denoyer \and Patrick Gallinari}
\institute{University Pierre et Marie Curie - UPMC, LIP6\\
Case 169 - 4 Place Jussieu - 75005 PARIS - FRANCE\\
{firstname.lastname@lip6.fr}
}

\maketitle
\begin{abstract}
We propose to model the text classification process as a sequential decision process. In this process, an agent learns to classify documents into topics while reading the document sentences sequentially and learns to stop as soon as enough information was read for deciding. The proposed algorithm is based on a modelisation of Text Classification as a Markov Decision Process and learns by using Reinforcement Learning. Experiments on four different classical mono-label corpora show that the proposed approach performs comparably to classical SVM approaches for large training sets, and better for small training sets. In addition, the model automatically adapts its reading process to the quantity of training information provided. 
\end{abstract}
 
\vspace{-1cm}
\section{Introduction}
\vspace{-0.3cm}
Text Classification (TC) is the act of taking a set of labeled text documents, learning a correlation between a document's contents and its corresponding labels, and then predicting the labels of a set of unlabeled test documents as best as possible. TC has been studied extensively, and is one of the older specialties of Information Retrieval. Classical statistical TC approaches are based on well-known machine learning models such as generative models --- Naive Bayes for example \cite{Lewis1994}\cite{Lewis1996} --- or discriminant models such as Support Vector Machines \cite{Joachims1998}. They mainly consider the \textit{bag of words} representation of a document (where the order of the words or sentences is lost) and try to compute a category score by looking at the entire document content. \textit{Linear} SVMs in particular --- especially for multi-label classification with many binary SVMs --- have been shown to work particularly well \cite{Dumais1998}. Some major drawbacks to these \textit{global} methods have been identified in the literature: 

\begin{itemize}
\item These methods take into consideration a document's entire word set in order to decide to which categories it belongs.  The underlying assumption is that the category information is homogeneously dispatched inside the document. This is well suited for corpora where documents are short, with little noise, so that global word frequencies can easily be correlated to topics. However, these methods will not be well suited in predicting the categories of large documents where the topic information is concentrated in only a few sentences.
\item  Additionally, for these methods to be applicable, the entire document must be known at the time of classification.  In cases where there is a cost associated with acquiring the textual information, methods that consider the entire document cannot be efficiently or reliably applied as we do not know at what point their classification decision is well-informed while considering only a subset of the document.
\end{itemize}

Considering these drawbacks, some attempts have been made to use the sequential nature of these documents for TC and similar problems such as passage classification. The earliest models developed especially for sequence processing extend Naive Bayes with Hidden Markov Models. Denoyer et al. \cite{Denoyer2001} propose an original model which aims at modeling a document as a sequence of irrelevant and relevant sections relative to a particular topic. 
In \cite{818051}, the authors propose a model based on recurrent Neural Networks for document routing. Other approaches have proposed to extend the use of linear SVMs to sequential data, mainly through the use of string kernels \cite{944799}. Finally, sequential models have been used for Information Extraction \cite{Leek97informationextraction,AminiZG00}, passage classification \cite{326445,1165775}, or the development of search engines \cite{Miller99bbnat,1793297}.

We propose a new model for Text Classification that is less affected by the aforementioned issues. Our approach models an agent that sequentially reads a text document while concurrently deciding to assign topic labels. This is modeled as a sequential process whose goal is to classify a document by focusing on its relevant sentences. The proposed model learns not only to classify a document into one or many classes, but also \textit{when} to label, and when to stop reading the document. This last point is very important because it means that the systems is able to learn to label a document with the correct categories as soon as possible, without reading the entire text. 

The contributions of this paper are three-fold:
\vspace{-6pt}
\begin{enumerate}
\item We propose a new type of sequential model for text classification based on the idea of sequentially reading sentences and assigning topics to a document.
\item Additionally, we propose an algorithm using Reinforcement Learning that learns to focus on relevant sentences in the document. This algorithm also learns when to stop reading a document so that the document is classified as soon as possible.  This characteristic can be useful for documents where sentence acquisition is expensive, such as large Web documents or conversational documents. 
\item We show that on popular text classification corpora our model outperforms classical TC methods for small training sets and is equivalent to a baseline SVM for larger training sets while only reading a small portion of the documents. The model also shows its ability to classify by reading only a few sentences when the classification problem is easy (large training sets) and to learn to read more sentences when the task is harder (small training sets).
\end{enumerate}
\vspace{-6pt}
This document is organized as follows: In Section \ref{sec:not-task}, we present an overview of our method. We formalize the algorithm as a Markov Decision Process in Section \ref{sec:tc-sdp} and detail the approach for both multi-label and mono-label TC. We then present the set of experiments made on four different text corpora in Section \ref{part:exp}.

\vspace{-0.4pt}
\section{Task Definition and General Principles of the Approach}
\vspace{-0.3pt}
 \label{sec:not-task}
	Let $\documents$ denote the set of all possible textual documents, and $\categories$ the set of $\nbcategories$ categories numbered from $1$ to $\nbcategories$. Each document $\doc{}$ in $\documents$ is associated with one or many\footnote{In this article, we consider both the mono-label classification task, where each document is associated with exactly one category, and the multi-label task where a document can be associated with several categories.} categories of $\mathcal{C}$. This label information is only known for a subset of documents $\documentstrain \subset \documents$ called training documents, composed of $\nbdocumentstrain$ documents denoted $\documentstrain=(\doc{1},...,\doc{\nbdocumentstrain})$. The labels of document $\doc{i}$ are given by a vector of scores $\scorecategorydocument{i}=(\scorecategorydocumentcategory{i}{1},...,\scorecategorydocumentcategory{i}{\nbcategories})$. We assume that:

\begin{equation}
		\scorecategorydocumentcategory{i}{k} = \begin{cases} 1 \text{ if } \doc{i} \text{ belongs to category } k \\
																												 0 \text{ otherwise}
																						\end{cases}.
\end{equation}
	The goal of TC is to compute, for each document $d$ in $\documents$, the corresponding score for each category.  The classification function $f_\theta$ with parameters $\theta$ is thus defined as : 
	\begin{equation}
		f_\theta : \begin{cases} \documents : [0;1]^\nbcategories \\ d \rightarrow \scorecategorydocument{d} \end{cases}.
	\end{equation}
	Learning the classifier consists in finding an optimal parameterization $\theta^*$ that reduces the mean loss such that:
    \begin{align}\label{eq:min-theta}
        \theta^* = \argmin_{\theta} {1 \over N_{train}} \sum_{i=1}^{N_{train}} L(f_\theta(d_i), y^{d_i}),
    \end{align}
    where $L$ is a loss function proportional to the classification error of $f_\theta(d_i)$.
    
\vspace{-0.5cm}
\subsection{Overview of the approach}
\label{ioverview}
This section aims to provide an intuitive overview of our approach.  The ideas presented here are formally presented in Section \ref{sec:tc-sdp}, and will only be described in a cursory manner below.
\vspace{-0.3cm}
\subsubsection{Inference} We propose to model the process of text classification as a sequential decision process.  In this process, our classifier reads a document sentence-by-sentence and can decide --- at each step of the reading process --- if the document belongs to one of the possible categories. 
This classifier can also chose to stop reading the document once it considers that the document has been correctly categorized. 

\begin{figure}[h]
\begin{center}
\hspace*{-0.25cm}
\includegraphics[width=1.12\linewidth]{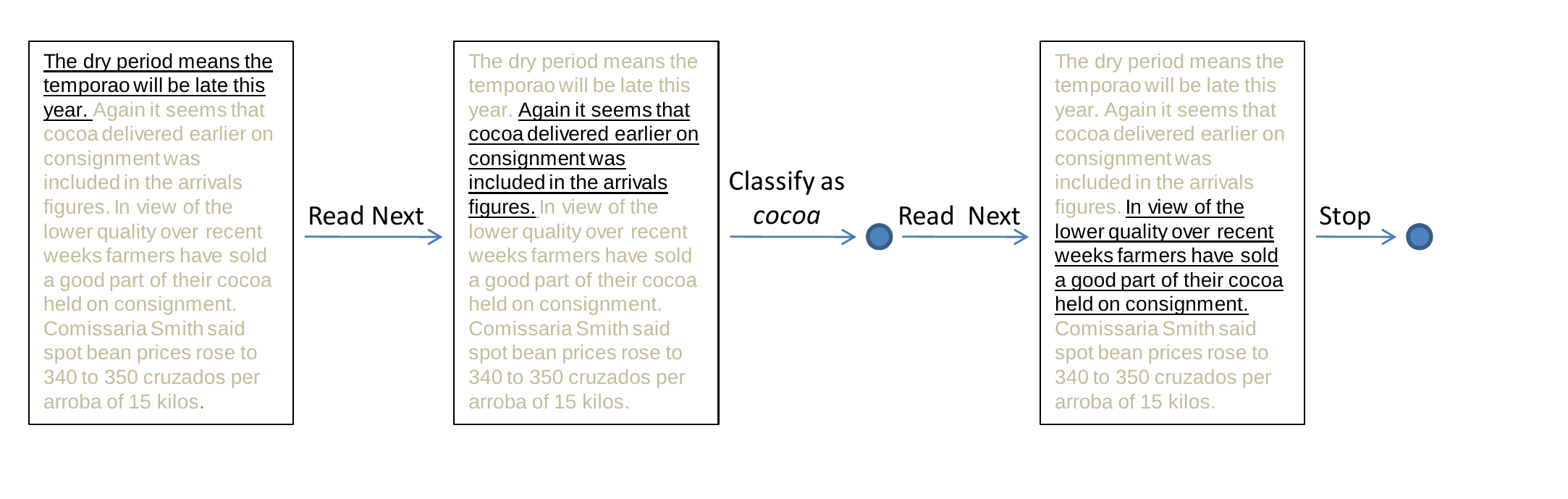}
\end{center}
\vspace{-1.0cm}
\caption{Inference on a document}
\label{fig:seqread}
\end{figure}

In the example described in Fig. \ref{fig:seqread}, the task is to classify a document composed of 4 sentences.  The documents starts off unclassified, and the classifier begins by reading the first sentence of the document. Because it considers that the first sentence does not contain enough information to reliably classify the document, the classifier decides to read the following sentence. Having now read the first two sentences, the classifier decides that it has enough information at hand to classify the document as \textit{cocoa}. 

\vspace{-0.5cm}The classifier now reads the third sentence and --- considering the information present in this sentence --- decides that the reading process is finished; the document is therefore classified in the \textit{cocoa} category.  

Had the document belonged to multiple classes, the classifier could have continued to assign other categories to the document as additional information was discovered.

In this example, the model took four \textbf{actions}: \textit{next}, \textit{classify as cocoa}, \textit{next} and then \textit{stop}.  The choice of each action was entirely dependent on the corresponding \textbf{state} of the reading process. The choice of actions given the state, such as those picked while classifying the example document above, is called the \textbf{policy} of the classifier. 
This policy --- denoted $\pi$ --- consists of a mapping of states to actions relative to a score.  This score is called a Q-value --- denoted $Q(s,a)$ --- and reflects the worth of choosing action $a$ during state $s$ of the process. 
Using the Q-value, the inference process can be seen as a \textbf{greedy process} which, for each timestep, chooses the best action $a^*$ defined as the action with the highest score w.r.t. $Q(s,a)$:
\begin{equation}
a^*=\argmax\limits_{a} Q(s,a).
\end{equation}
\vspace{-0.3cm}
\subsubsection{Training}The learning process consists in computing a $Q$-$function$\footnote{The \textit{Q-function} is an approximation of $Q(s,a)$.} which minimizes the classification loss (as in equation \eqref{eq:min-theta}) of the documents in the training set. The learning procedure uses a monte-carlo approach to find a set of \textit{good} and \textit{bad} actions relative to each state.  \textit{Good} actions are actions that result in a small classification loss for a document. The \textit{good} and \textit{bad} actions are then learned by a statistical classifier, such as an SVM. 

\begin{figure}[h]
\begin{center}
\includegraphics[width=1.01\linewidth]{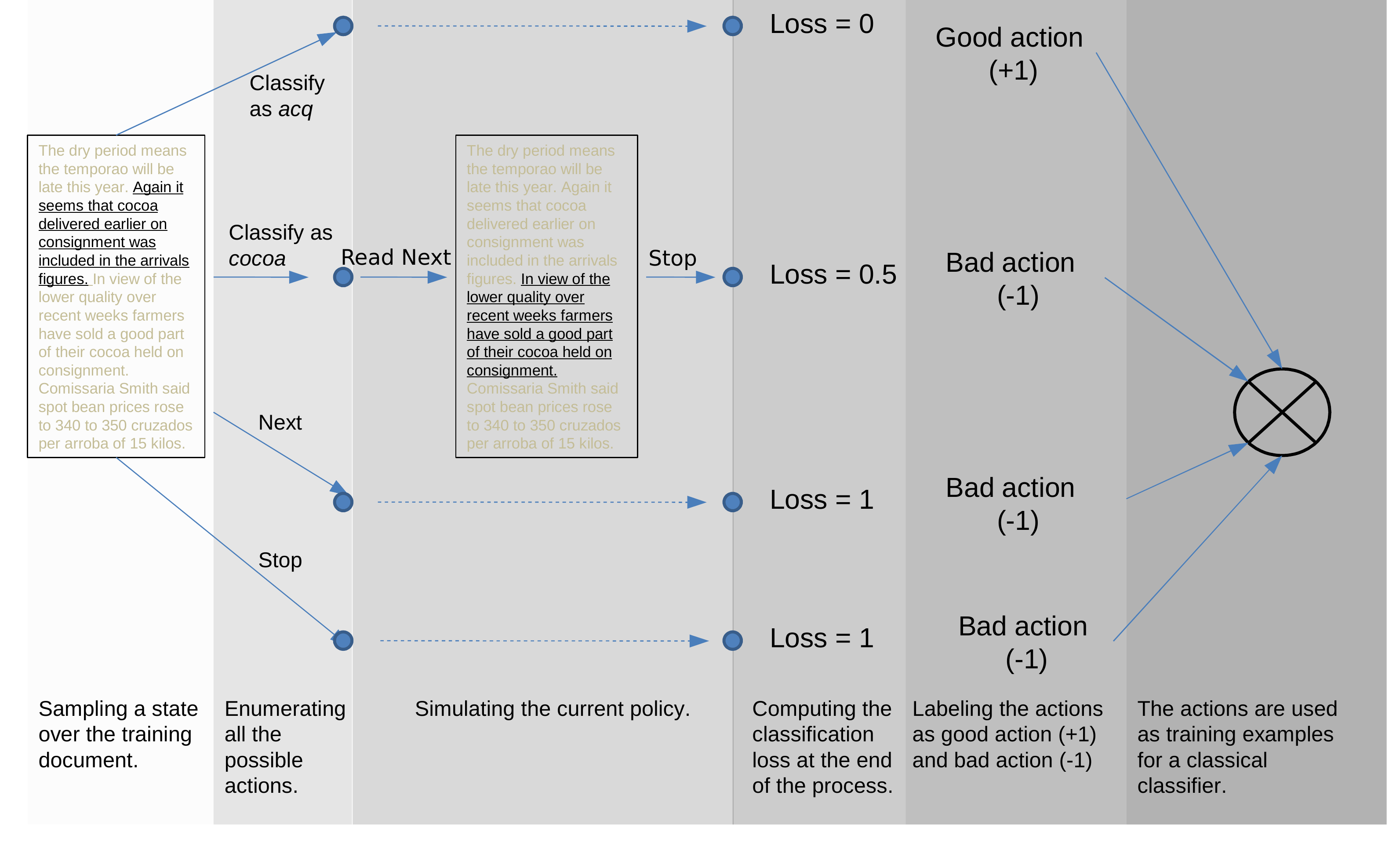}
\end{center}
\vspace{-0.5cm}
\caption{Learning the sequential model. The different steps of one learning iteration are illustrated from left to right on a single training document.}
\label{fig:seqlearn}
\end{figure}

An example of the training procedure on the same example document as above is illustrated in Fig \ref{fig:seqlearn}.  To begin with, a random state of the classification process is picked. Then, for each action possible in that state, the current policy is run until it stops and the final classification loss is computed. 
The training algorithm then builds a set of \textit{good} actions --- the actions for which the simulation obtains the minimum loss value --- and a set of remaining \textit{bad} actions. This is repeated on many different states and training documents until, at last, the model learns a classifier able to discriminate between \textit{good} and \textit{bad} actions relative to the current state. 

\vspace{-0.8cm}
\subsection{Preliminaries}

We have presented the principles of our approach and given an intuitive description of the inference and learning procedures.  We will now formalize this algorithm as a Markov Decision Process (MDP) for which an optimal policy is found via Reinforcement Learning. Note that we will only go over notations pertinent to our approach, and that this section lacks many MDP or Reinforcement Learning definitions that are not necessary for our explanation.
\vspace{-0.3cm}
\subsubsection{Markov Decision Process}
	A Markov Decision Process (MDP) is a mathematical formalism to
model sequential decision processes.  We only consider deterministic MDPs, defined by a 4-tuple: $(\states,\actions,\transitions,\reward)$. Here, $\states$ is the set of possible states, $\actions$ is the is the set of possible actions, and $\transitions : S \times A
\rightarrow S$ is the state transition function such that $\transition{s}{a}\rightarrow s'$ (this symbolizes the system moving from state $s$ to state $s'$ by applying action $a$).  The reward, $\reward : S \times A \rightarrow \mathbb{R}$, is a value that reflects the quality of taking action $a$ in state $s$ relative to the agent's ultimate goal.  We will use $\possibleactions{s} \subseteq \actions$ to refer to the set of possible actions available to an agent in a particular state $s$.

An agent interacts with the MDP by starting off in a state $s \in \states$.  The agent then
chooses an action $a \in \possibleactions{s}$ which moves it to a new state $s'$ by applying the transition $\transition{s}{a}$. The agent obtains a reward $\rewardsa{s}{a}$ and then continues the process until it reaches a terminal state $s_{final}$ where the set of possible actions is empty i.e $\possibleactions{s_{final}}=\emptyset$. 
\vspace{-0.3cm}
\subsubsection{Reinforcement Learning}
Let us define $\pi : S \rightarrow A$, a stochastic policy such that $\forall a \in A(s)$, $\pi(s) = a$ with probability $P(a|s)$. The goal of RL is to find an optimal policy $\pi^*$ that maximizes the cumulative reward obtain by an agent. We consider here the finite-horizon context for which the cumulative reward corresponds to the sum of the reward obtained at each step by the system, following the policy $\pi$.  The goal of Reinforcement Learning is to find an optimal policy denoted $\pi^*$  which maximizes the cumulative reward obtained for all the states of the process i.e.:
\begin{equation}
\pi^* = \argmax_\pi \sum\limits_{s_0 \in \states} \mathbb{E}_{\pi}[\sum\limits_{t=0}^{T}r(s_t,a_t)].
\end{equation}

Many algorithms have been developed for finding such a policy, depending on the assumptions made on the structure of the MDP, the nature of the states (discrete or continuous), etc.  In many approaches, a policy $\pi$ is defined through the use of a $Q$-$function$ which reflects how much reward one can expect by taking action $a$ on state $s$. With such a function, the policy $\pi$ is defined as:
\begin{equation}
\pi = \argmax\limits_{a \in \mathcal{A}(s)} Q(s,a).
\end{equation}
In such a case, the learning problem consists in finding the optimal value $Q^*$ which results in the optimal policy $\pi^*$.

Due to the very large number of states we are facing in our approach, we consider the Approximated Reinforcement Learning context where the $Q$ function is approximated by a parameterized function $Q_\theta(s,a)$, where $\theta$ is a set of parameters such that:
\begin{equation}
Q_\theta(s,a) = <\theta, \Phi(s,a)>,
\end{equation}
where $<\cdot,\cdot>$ denotes the dot product and $\Phi(s,a)$ is a feature vector representing the state-action pair $(s,a)$.  The learning problem consists in finding the optimal parameters $\theta^*$ that results in an optimal policy:
\begin{equation}
\pi^* = \argmax\limits_{a \in \mathcal{A}(s)} <\theta^*,\Phi(s,a)>.
\end{equation}

\vspace{-0.5cm}
\section{Text Classification as a Sequential Decision Problem}
\label{sec:tc-sdp}
\vspace{-0.3cm}
Formally, we consider that a document $\doc{}$ is composed of a sequence of sentences such that $\doc{}=(\sentence{\doc{}}{1},\hdots,\sentence{\doc{}}{\lengthdoc{\doc{}}})$, where $\sentence{\doc{}}{i}$ is the i-th sentence of the document and $\lengthdoc{\doc{}}$ is the total number of sentences making up the document.  Each sentence $\sentence{\doc{}}{i}$ has a corresponding feature vector --- a normalized tf-idf vector in our case --- that describes its content.
\vspace{-0.4cm}
\subsection{MDP for Multi-Label Text Classification}\label{sec:sdp-mdp}
\vspace{-0.3cm}
 We can describe our sequential decision problem using an MDP.  Below, we describe the MDP for the multilabel classification problem, of which monolabel classification is just a specific instance:
\begin{itemize}
\item Each state $s$ is a triplet $(\doc{},p,\estimatedscorecategorydocumentcategory{})$ such that:
\begin{itemize}
\item $\doc{}$ is the document the agent is currently reading.
\item $p \in [1,\lengthdoc{\doc{}}]$ corresponds to the current sentence being read; this implies that $\sentence{\doc{}}{1}$ to $\sentence{\doc{}}{p-1}$ have already been read.
\item $\estimatedscorecategorydocumentcategory{}$ is the set of currently assigned categories --- categories previously assigned by the agent during the current reading process --- where $\estimatedscorecategorydocumentcategory{k} = 1$ iff the document has been assigned to category $k$ during the reading process, $0$ otherwise.
\end{itemize}

\item The set of actions $\possibleactions{s}$ is composed of:
\begin{itemize}
\item One or many \textbf{classification} actions denoted \textit{classify as $k$} for each category $k$ where $\estimatedscorecategorydocumentcategory{k} = 0$.  These actions correspond to assigning document $\doc{}$ to category $k$.
\item A \textbf{next sentence} action denoted \textit{next} which corresponds to reading the next sentence of the document.
\item A \textbf{stop action} denoted \textit{stop} which corresponds to finishing the reading process.
\end{itemize}

\item The set of transitions $\transition{s}{a}$ act such that:
\begin{itemize}
\item $\transition{s}{classify as k}$ sets $\estimatedscorecategorydocumentcategory{k} \leftarrow 1$.
\item $\transition{s}{next}$ sets $p \leftarrow p+1$.
\item $\transition{s}{stop}$ halts the decision process.
\end{itemize}

\item The reward $\reward(s,a)$ is defined as:
\begin{align}
r(s,a) = \left\{
\begin{array}
{lll}
F_1(y,\hat{y})&\mbox{ if } a \textrm{ is a \textit{stop} action}\\
0&\mbox{otherwise}
\end{array}
\right .,
\end{align}
where $y$ is the real vector of categories for $d$ and $\hat{y}$ is the predicted vector of categories at the end of the classification process. The $F_1$ score of a single document is defined as:
\begin{align}
F_1(y,\hat{y}) &= 2 \cdot \frac{p(y,\hat{y}) \cdot r(y,\hat{y})}{p(y,\hat{y}) + r(y,\hat{y})}\\
&\text{ with }\\
 p(y,\hat{y}) = \sum\limits_{k=0}^C \mathds{1} (\hat{y_k} = y_k) / \sum\limits_{k=0}^C \hat{y_k} &\text{ and }r(y,\hat{y}) = \sum\limits_{k=0}^C \mathds{1} (\hat{y_k} = y_k) / \sum_{k=0}^C y_k.
\end{align}
\end{itemize}

%

\subsubsection{MDP for Mono-Label Text Classification}
In mono-label classification, we restrict the set of possible actions. The \textit{classify as k} action leads to a stopping state such that $A(s) = \{\textit{stop}\}$.  This brings the episode to an end after the attribution of a single label. Note that in the case of a mono-label system --- where only one category can be assigned to a document --- the reward corresponds to a classical accuracy measure: $1$ if the chosen category is correct, and $0$ otherwise.
\vspace{-0.4cm}
\subsection{Features over states}
\vspace{-0.3cm}
We must now define a feature function which provides a vector representation of a state-action pair $(s,a)$.  The purpose of this vector is to be able to present $(s,a)$ to a statistical classifier to know whether it is \textit{good} or \textit{bad}.  Comparing the scores of various $(s,a)$ pairs for a given state $s$ allows us to choose the best action for that state. 

Classical text classification methods only represent documents by a global --- and usually tf-idf weighted --- vector.  We choose, however, to include not only a global representation of the sentences read so far, but also a local component corresponding to the most recently read sentence. Moreover, while in state $s$, a document may have been already assigned to a set of categories; the global feature vector $\Phi(s,a)$ must describe all this information. The vector representation of a state $s$ is thus defined as $\Phi(s)$:
\begin{equation}
\Phi(s) = \left( \frac{ \sum\limits_{i=1}^{p} \sentence{\doc{}}{i} }{p} \text{          }  \sentence{\doc{}}{p} \text{          } \hat{y_0} \hdots  \hat{y_C} \right).
\end{equation}
$\Phi(s)$ is the concatenation of a set of sub-vectors describing: the mean of the feature vectors of the read sentences, the feature vector of the last sentence, and the set of already assigned categories.

In order to include the action information, we use the block-vector trick introduced by \cite{Har-Peled2002} which consists in projecting $\Phi(s)$ into a higher dimensional space such that:
\begin{equation}
\Phi(s,a) = \left( 0 \hdots \phi(s) \hdots 0 \right).
\end{equation}
The position of $\Phi(s)$ inside the global vector $\Phi(s,a)$ is dependent on action $a$. This results in a very high dimensional space which is easier to classify in with a linear model.

\vspace{-0.4cm}
\subsection{Learning and Finding the optimal classification policy}
\vspace{-0.3cm}
In order to find the best classification policy, we used a recent Reinforcement Learning algorithm called \textit{Approximate Policy Iteration with Rollouts}. In brief, this method uses a monte-carlo approach to evaluate the quality of all the possible actions amongst some random sampled states, and then learns a classifier whose goal is to discriminate between the \textit{good} and \textit{bad} actions relative to each state. Due to a lack of space, we do not detail the learning procedure here and refer to the paper by Lagoudakis et al \cite{Lagoudakis2003}. An intuitive description of the procedure is given in Section \ref{ioverview}.

\vspace{-0.3cm}
\section{Experimental Results}
\label{part:exp}
\vspace{-0.3cm}
We have applied our model on four different popular datasets: three mono-label and one multi-label.  All datasets were pre-processed in the same manner: all punctuation except for periods were removed, SMART stop-words\cite{Salton:71} and words less than three characters long were removed, and all words were stemmed with Porter stemming. Baseline evaluations were performed with libSVM\cite{Chang2001} on normalized tf-idf weighted vectorial representations of each document as has been done in \cite{Joachims1998}.  Published performance benchmarks can be found in \cite{Sebastiani2002} and \cite{Kumar2010}.  The datasets are:
\vspace{-0.3cm}
\begin{itemize}
\item The Reuters-21578\footnote{http://web.ist.utl.pt/\%7Eacardoso/datasets/} dataset which provides two corpora:
\begin{itemize}
\item The \textbf{Reuters8} corpus, a mono-label corpus composed of the 8 largest categories.
\item The \textbf{Reuters10} corpus, a multi-label corpus composed of the 10 largest categories.
\end{itemize}
\item The WebKB\footnote{http://www.cs.cmu.edu/afs/cs.cmu.edu/project/theo-20/www/data/}\cite{Craven1998} dataset is a mono-label corpus composed of Web pages dispatched into 4 different categories.
\item The 20 Newsgroups\footnote{http://people.csail.mit.edu/jrennie/20Newsgroups/} (20NG) dataset is a mono-label corpus of news composed of 20 classes. \end{itemize}
\vspace{-0.9    cm}
\begin{table}
\begin{center}
\begin{tabular}{|c||l|l|l|l|} \hline
Corpus & Nb of documents & Nb of categories & Nb of sentences by doc. & Task \\ \hline \hline
R8 & 7678 & 8 & 8.19 & Mono-label \\
R10 & 12 902 & 10 & 9.13 & Multi-label \\
Newsgroup & 18 846 & 20 &  22.34 & Mono-label \\
WebKB & 4 177 & 4 & 42.36 & Mono-label \\ \hline 
\end{tabular}
\end{center}
\caption{Corpora statistics.}
\vspace{-0.9cm}
\label{stats}
\end{table}

\vspace{-0.7cm}
\subsection{Evaluation Protocol}
\vspace{-0.3cm}
Many classification systems are \textit{soft classification} systems that compute a score for each possible category-document pair.  Our system is a \textit{hard classification} system that assigns a document to one or many categories, with a score of either 1 or 0.  The evaluation measures used in the litterature, such as the \textit{breakeven} point, are not suitable for \textit{hard classification} models and cannot be used to evaluate and compare our approach with other methods.  We have therefore chosen to use the \textit{micro-F1} and \textit{macro-F1} measures. These measures correspond to a classical $F_1$ score computed for each category and averaged over the categories. The \textit{macro-F1} measure does not take into account the size of the categories, whereas the \textit{micro-F1} average is weighted by the size of each category. We averaged the different models' performances on various train/test splits that were randomly generated from the original dataset.  We used the same approach both for evaluating our approach and the baseline approaches to be able to compare our results properly. For each training size, the performance of the models were averaged over 5 runs. The hyper-parameters of the SVM and the hyper-parameters of the RL-based approach were manually tuned. What we present here are the best results obtained over the various parameter choices we tested. For the RL approach, each policy was learned on 10,000 randomly generated states, with 1 rollout per state, using a random initial policy. It is important to note that, in a practical sense, the RL method is not much more complicated to tune than a classical SVM since it is rather robust regarding the values of the hyper-parameters.
\vspace{-0.1cm}
\subsection{Experimental Results}
\begin{figure}
\begin{tabular}{cc}
 \hspace{-2cm} \includegraphics[width=0.7\linewidth, trim = 0 30mm 0 30mm, clip=true]{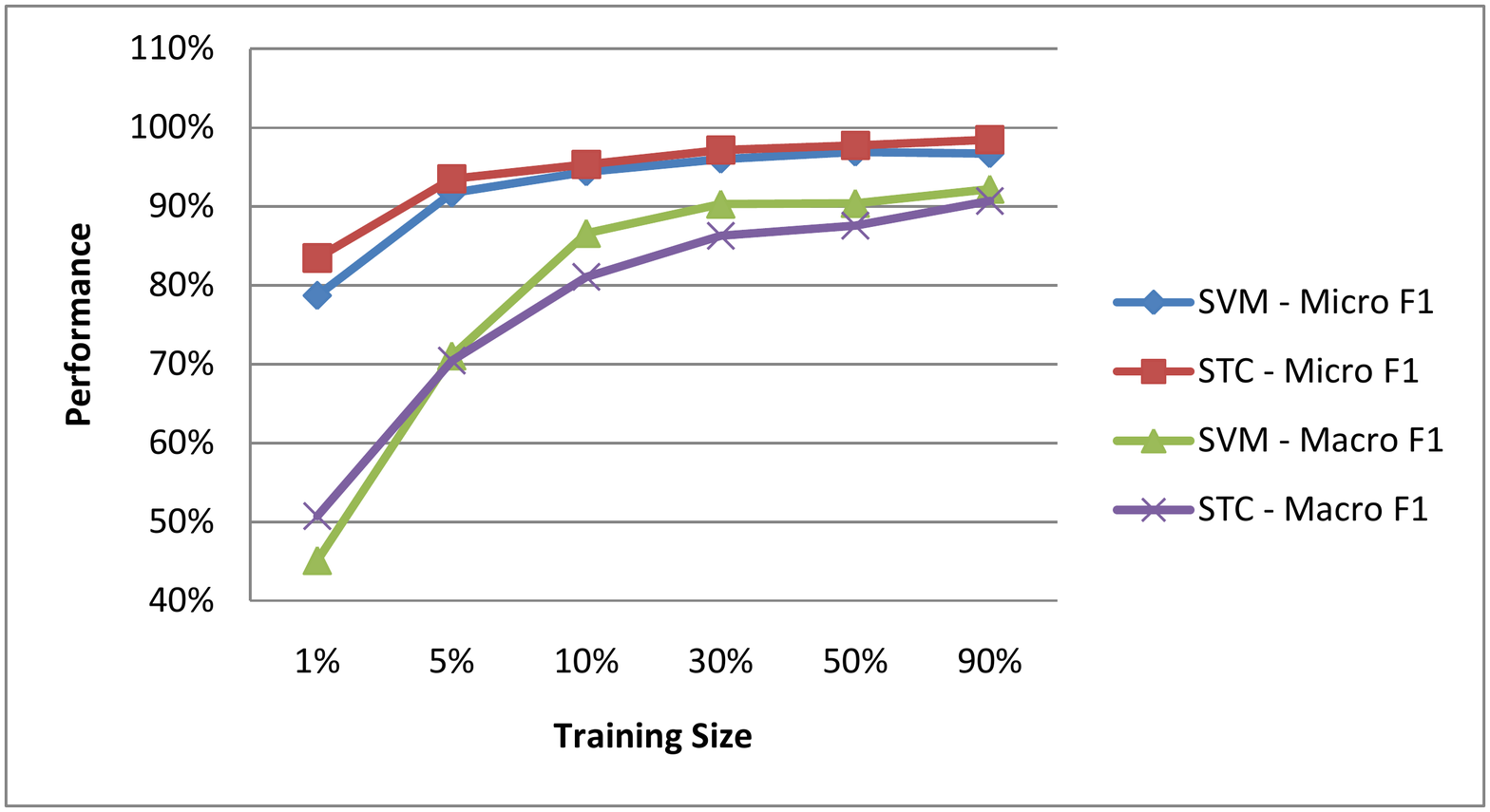} & \hspace{-1cm} \includegraphics[width=0.7\linewidth, trim = 0 30mm 0 30mm, clip=true]{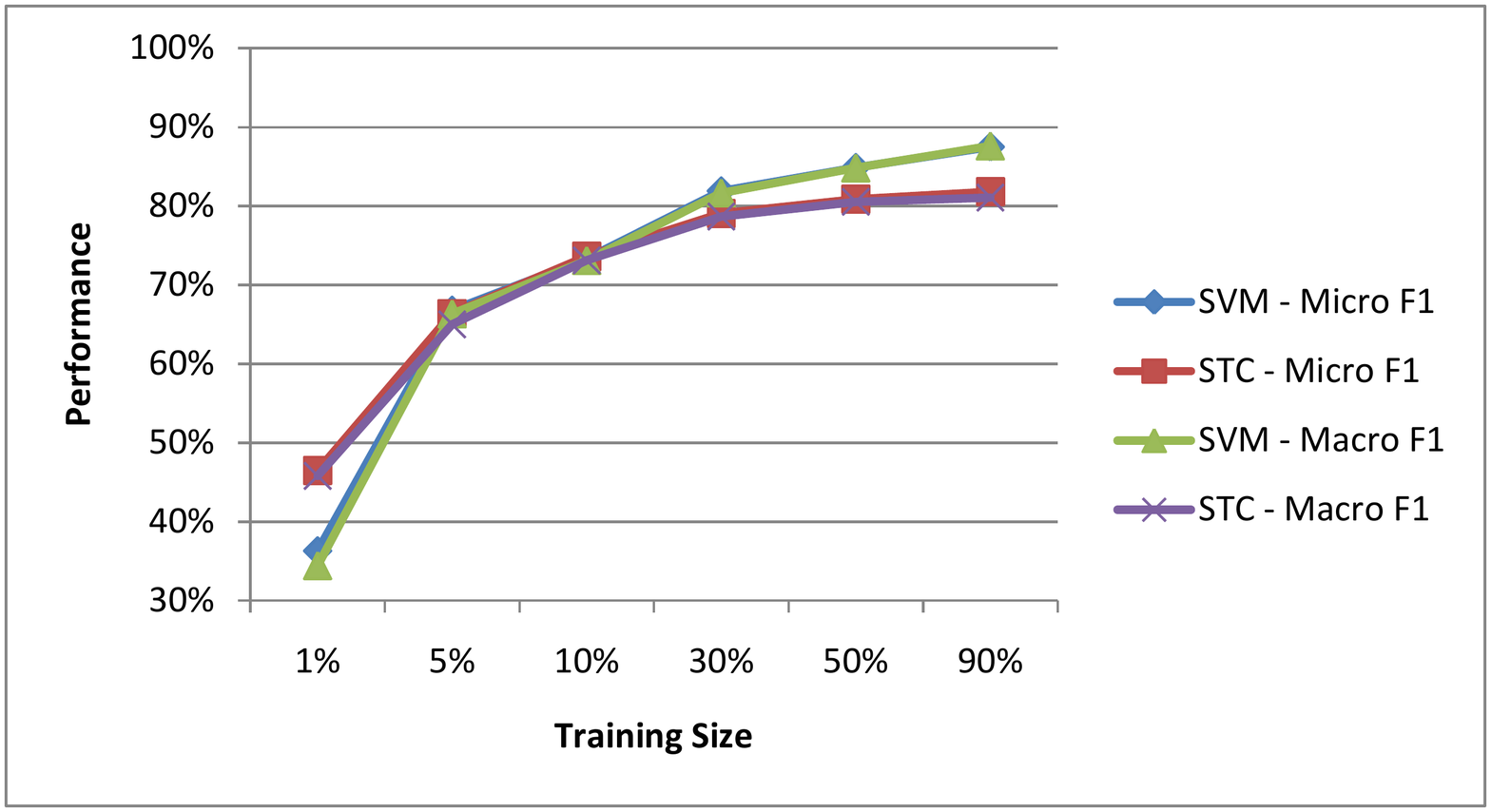}
\end{tabular}
\caption{Performances over the R8 Corpus (left) and NewsGroup Corpus (right)}
\label{fig:mono-1}
\end{figure}

\begin{figure}
\begin{tabular}{cc}
\hspace{-2cm} \includegraphics[width=0.7\linewidth, trim = 0 30mm 0 30mm, clip=true]{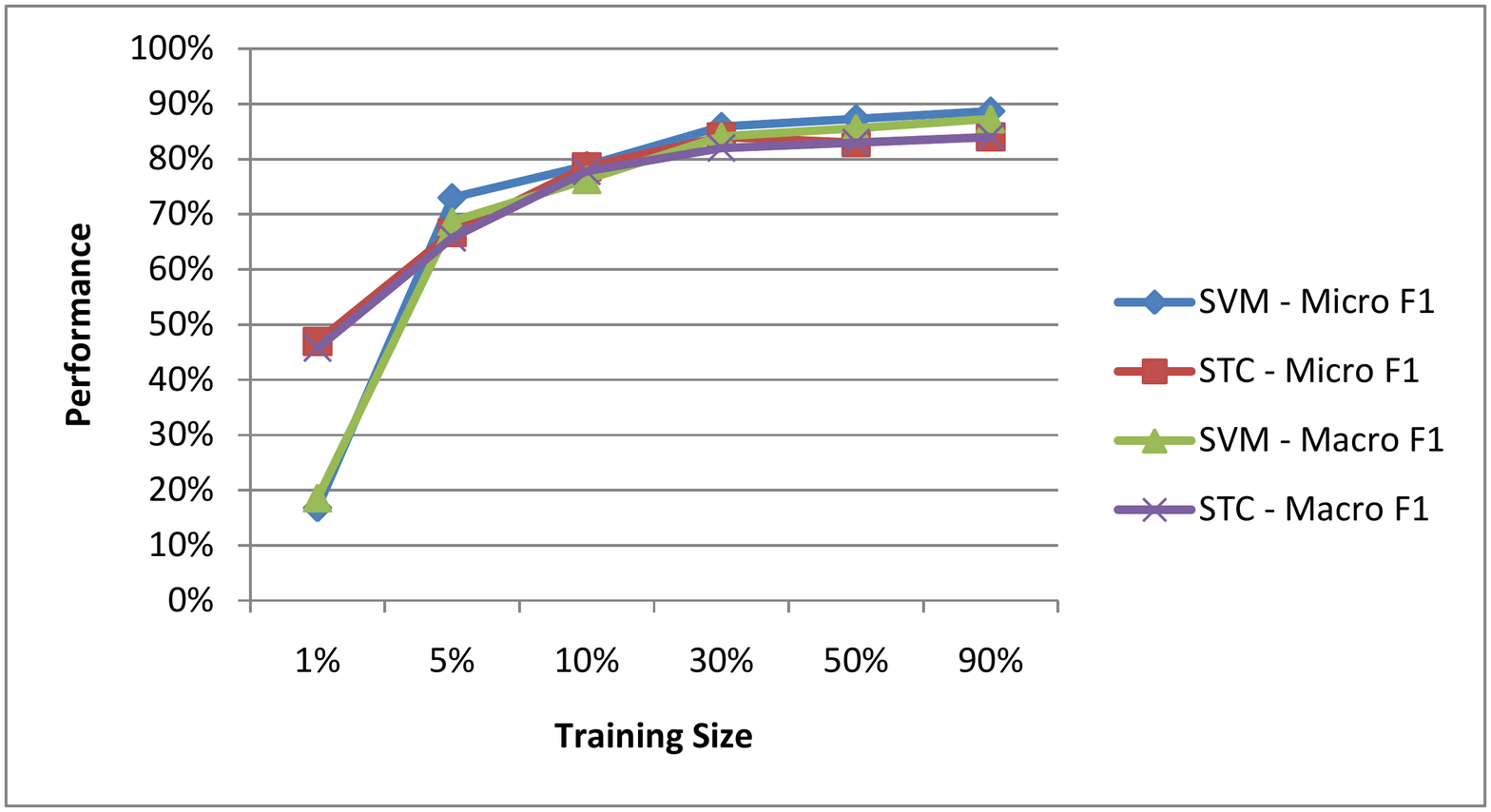} & \hspace{-1cm} \includegraphics[width=0.7\linewidth, trim = 0 30mm 0 30mm, clip=true]{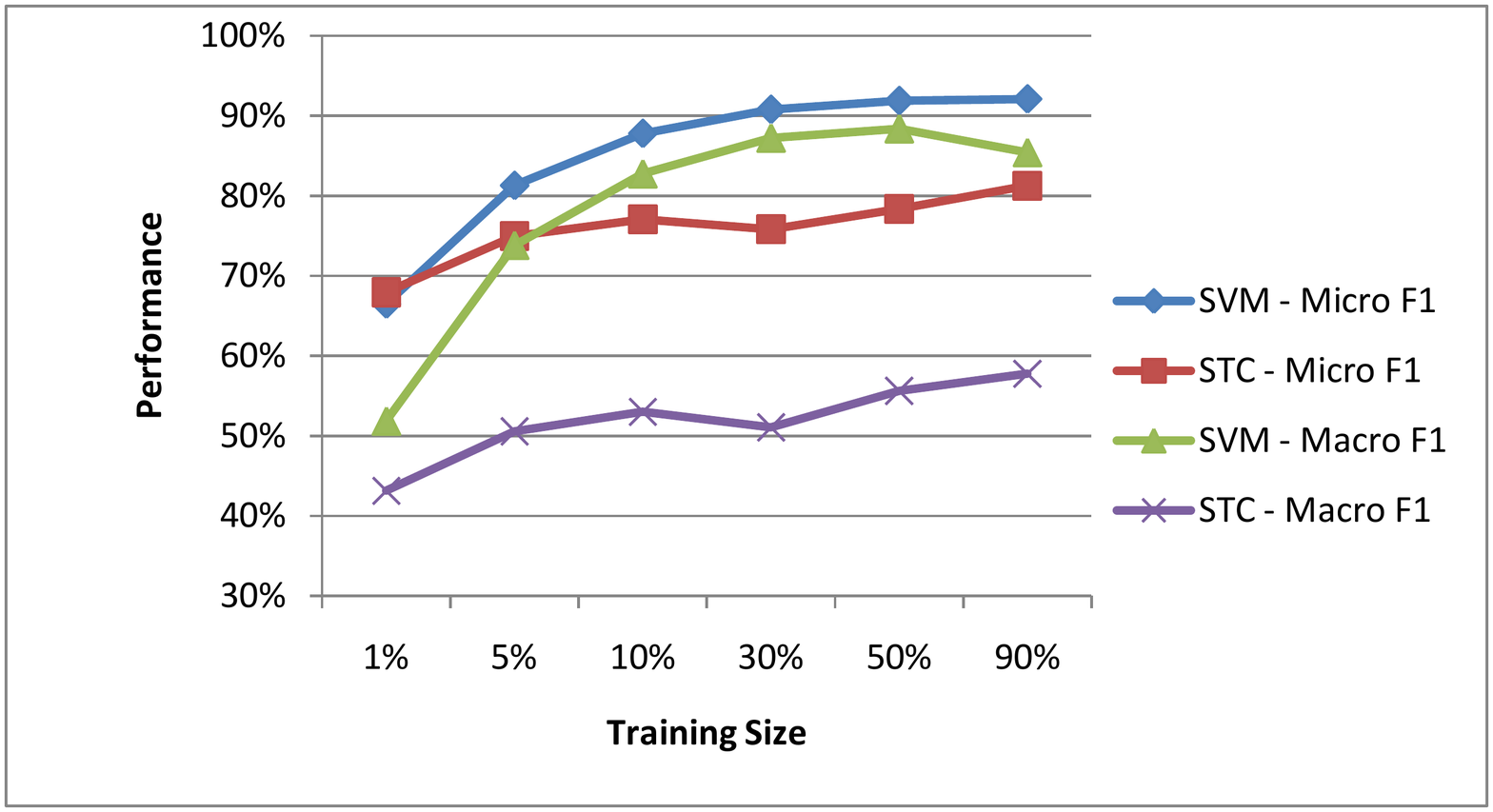}
\end{tabular}
\caption{Performances over the WebKB Corpus (left) and R10 Corpus (right)}
\label{fig:mono-multi}
\end{figure}

\vspace{-0.25cm}
Our performance figures use SVM to denote baseline Support Vector Machine performance, and STC (Sequential Text Classification) to denote our approach. In the case of the mono-label experiments (Figure \ref{fig:mono-1} and \ref{fig:mono-multi}-left), performance of both the SVM method and our method are comparable. It is important to note, however, that in the case of small training sizes (1\%, 5\%), the STC approach outperforms SVM by 1-10\% depending on the corpus. For example, on the R8 dataset we can see that for both \textit{F1} scores, STC is better by $\sim5\%$ with a training size of 1\%.  This is also visible with the NewsGroup dataset, where STC is better by 10\% for both metrics using a 1\% training set. This shows that STC is particularly advantageous with small training sets. 
 
\begin{figure}
\begin{tabular}{cc}
 \hspace{-2cm} \includegraphics[width=0.7\linewidth, trim = 0 30mm 0 30mm, clip=true]{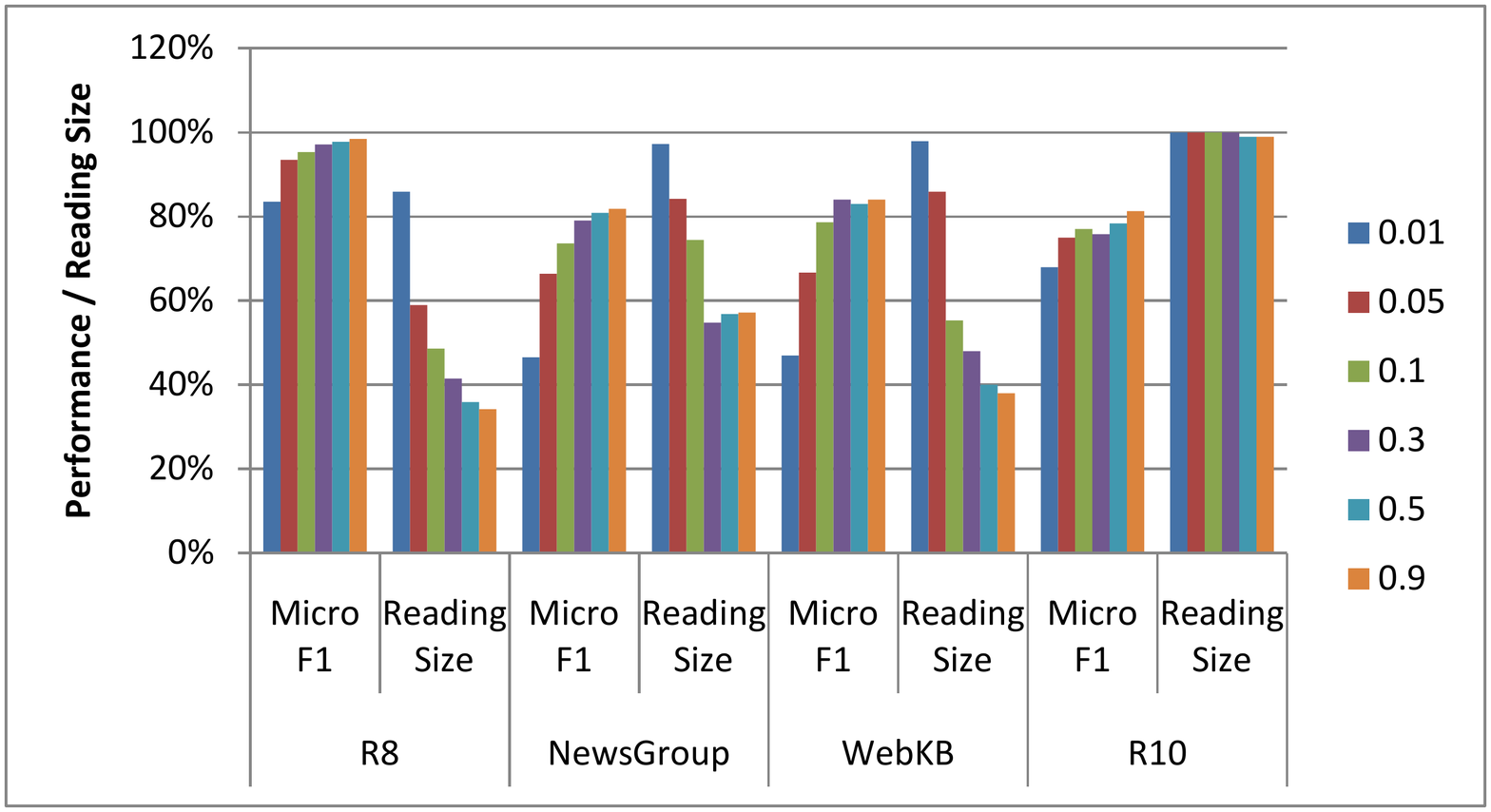} & \hspace{-1cm} \includegraphics[width=0.7\linewidth, trim = 0 30mm 0 30mm, clip=true]{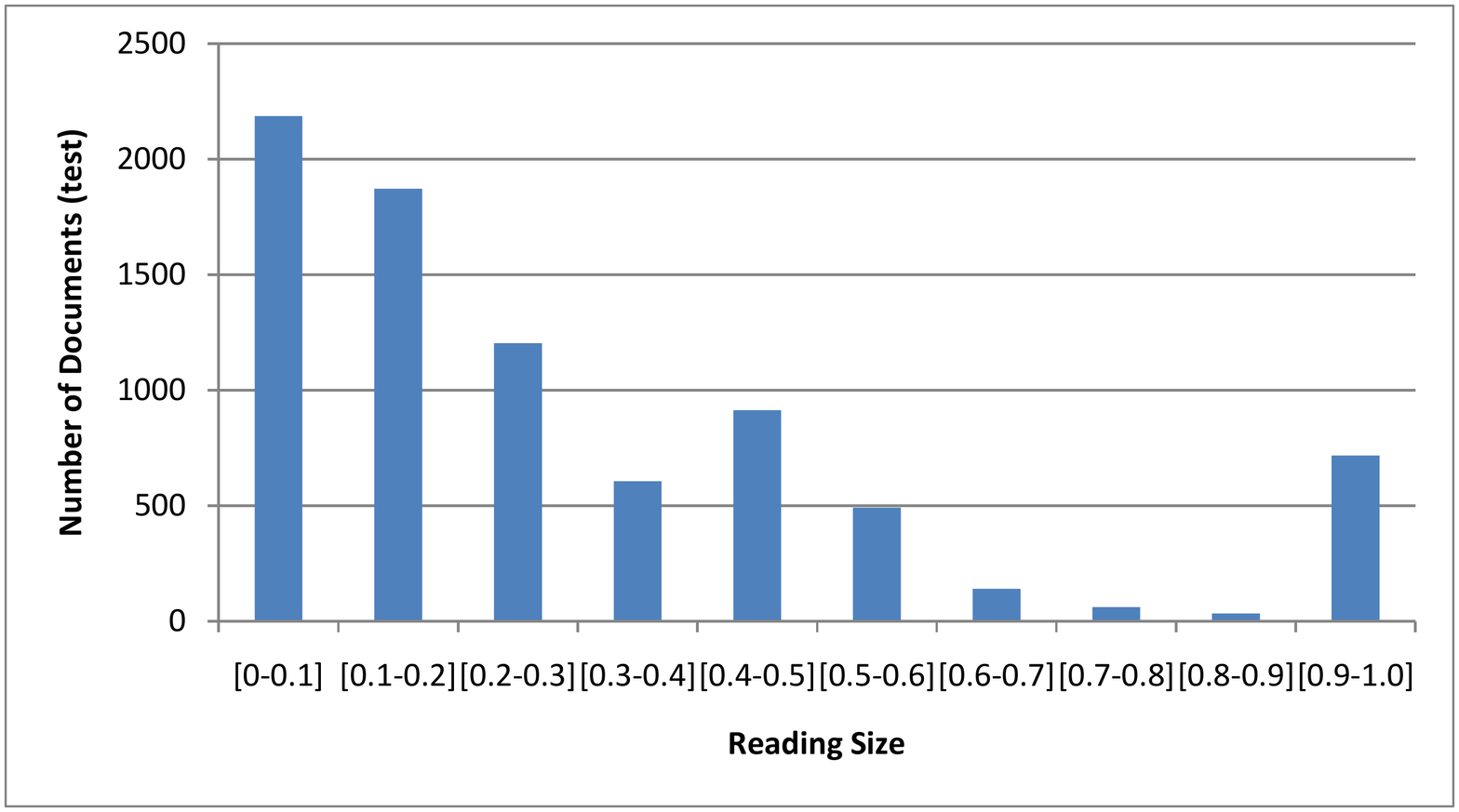}
\end{tabular}
\caption{Overview of the Reading Sizes for all the corpora (left). Number of documents and Reading Sizes on R8 with 30\% of documents as a training set (right). }
\label{fig:read-perfs}
\end{figure}

The reading process' behaviour is explored in Figure \ref{fig:read-perfs}. Here, \textit{Reading Size} corresponds to the mean percentage of sentences read for each document\footnote{If $l_i$ is the number of sentences in document $i$ read during the classification process, and $n_i$ is the total number of sentences in this document. Let $N$ be the number of test documents, then the reading size value is $\frac{1}{N}\sum\limits_i \frac{l_i}{n_i}$.}. We can see that \textit{Reading Size} decreases as the training size gets bigger for mono-label corpora. This is due to the fact that the smaller training sizes are harder to learn, and therefore the agent needs more information to properly label documents. In the right-hand side of Figure \ref{fig:read-perfs}, we can see a histogram of number of documents grouped by \textit{Reading Size}. We notice that although there is a mean \textit{Reading Size} of 41\%, most of the documents are barely read, with a few outliers that are read until the end.  The agent is clearly capable of choosing to read more or less of the document depending on its content.

In the case of multi-label classification, results are quite different. First, we see that for the R10 corpus, our model's performance is lower than the baseline on large training sets. Moreover, the multi-label model reads all the sentences of the document during the classification process. This behaviour seems normal because when dealing with multi-label documents, one cannot be sure that the remaining sentences will not contain relevant information pertaining to a particular topic. We hypothesize that the lower performances are due to the much larger action space in the multi-label problem, and the fact that we are learning a single model for all classes instead of one independent models per class.

\vspace{-0.5cm}
\section{Conclusions}
\vspace{-0.4cm}
We have presented a new model that learns to classify by sequentially reading the sentences of a document, and which labels this document as soon as it has collected enough information. This method shows some interesting properties on different datasets. Particularly in mono-label TC, the model automatically learns to read only a small part of the documents when the training set is large, and the whole documents when the training set is small. It is thus able to adapt its behaviour to the difficulty of the classification task, which results in obtaining faster systems for easier problems. The performances obtained are close to the performance of a baseline SVM model for large training sets, and better for small training sets. 

\vspace{-0.2cm}
This work opens many new perspectives in the Text Classification domain. Particularly, it is possible to imagine some additional MDP actions for the classification agent allowing the agent to parse the document in a more complex manner. For example, this idea can be extended to learn to classify XML documents reading only the relevant parts.
\vspace{-0.55cm}
\section*{Acknowledgments}
\vspace{-0.35cm}
This work was partially supported by the French National
Agency of Research (Lampada  ANR-09-EMER-007).
\bibliographystyle{IEEEtran}
\vspace{-0.55cm}
\bibliography{ECIR2011}
\end{document}